\documentclass[final,10pt,twocolumn]{IEEEtran}
\usepackage{lineno}
\usepackage{graphicx}
\usepackage{amsmath}
\usepackage{amssymb}
\usepackage{algorithmic}
\usepackage[ruled]{algorithm2e}
\usepackage{array}
\usepackage{booktabs}

\newcommand{\tabincell}[2]{\begin{tabular}{@{}#1@{}}#2\end{tabular}}

\ifCLASSINFOpdf
  \DeclareGraphicsExtensions{.pdf,.jpeg,.png}
\else
  \DeclareGraphicsExtensions{.eps}
\fi

\begin{document}

\title{Complex Background Subtraction by Pursuing Dynamic Spatio-Temporal Models}

%

\author{Liang~Lin,
        Yuanlu~Xu,~
        Xiaodan~Liang,~
        Jianhuang Lai~
\thanks{This work was supported by the Hi-Tech Research and Development (863) Program of China (no. 2012AA011504), National Natural Science Foundation of
China (no. 61173082, no. 61173084), Guangdong Science and Technology Program (no. 2012B031500006), Guangdong Natural Science Foundation (no. S2013050014548), Special Project on Integration of Industry, Education and Research of Guangdong Province (no. 2012B091000101), and Fundamental Research Funds for the Central Universities (no. 13lgjc26).\protect\\Copyright (c) 2014 IEEE. Personal use of this material is permitted.
However, permission to use this material for any other purposes must be obtained from the IEEE by sending an email to pubs-permissions@ieee.org.}\IEEEcompsocitemizethanks{\IEEEcompsocthanksitem L. Lin is with the Key Laboratory of Machine Intelligence and Advanced Computing (Sun Yat-Sen University), Ministry of Education, China, with the School of Advanced Computing, Sun Yat-Sen University, Guangzhou 510006, P. R. China, and with the SYSU-CMU Shunde International Joint Research Institute, Shunde, China. E-mail: linliang@ieee.org

\IEEEcompsocthanksitem Y. Xu, X. Liang and J. Lai are with the School of Information Science and Technology, Sun Yat-Sen University, Guangzhou 510006, P. R. China.}}

\markboth{IEEE TRANSACTIONS ON IMAGE PROCESSING, 2014.}%
{L. Lin \MakeLowercase{\textit{et al.}}: Complex Background Subtraction by Pursuing Dynamic Spatio-temporal Models}

\maketitle

\begin{abstract}

Although it has been widely discussed in video surveillance, background subtraction is still an open problem in the context of complex scenarios, \emph{e.g.}, dynamic backgrounds, illumination variations, and indistinct foreground objects. To address these challenges, we propose an effective background subtraction method by learning and maintaining an array of dynamic texture models within the spatio-temporal representations. At any location of the scene, we extract a sequence of regular video bricks, \emph{i.e.} video volumes spanning over both spatial and  temporal domain.  The background modeling is thus posed as pursuing subspaces within the video bricks while adapting the scene variations.  For each sequence of video bricks, we pursue the subspace by employing the ARMA (Auto Regressive Moving Average) Model that jointly characterizes the appearance consistency and temporal coherence of the observations.  During online processing,  we incrementally update the subspaces to cope with disturbances from foreground objects and scene changes. In the experiments, we validate the proposed method in several complex scenarios, and show superior performances over other state-of-the-art approaches of background subtraction. The empirical studies of parameter setting and component analysis are presented as well.

\end{abstract}





\IEEEpeerreviewmaketitle


\section{Introduction}

Background subtraction (also referred as foreground extraction) has been extensively studied in decades\cite{GGMM,bouwmans2008background,Maddalena08,ICA09,ChangJC12,liu2011Surveillance}, yet it still remains open in real surveillance applications due to the following challenges:

{\small \textbullet} \; Dynamic backgrounds. A scene environment is not always static but sometimes highly dynamic, \emph{e.g.}, rippling water, heavy rain and camera jitter.

{\small \textbullet} \; Lighting and illumination variations, particularly with sudden changes.

{\small \textbullet} \; Indistinct foreground objects having similar appearances with surrounding backgrounds.

\begin{figure}[ptb]
\begin{center}
\includegraphics[width=3.4in]{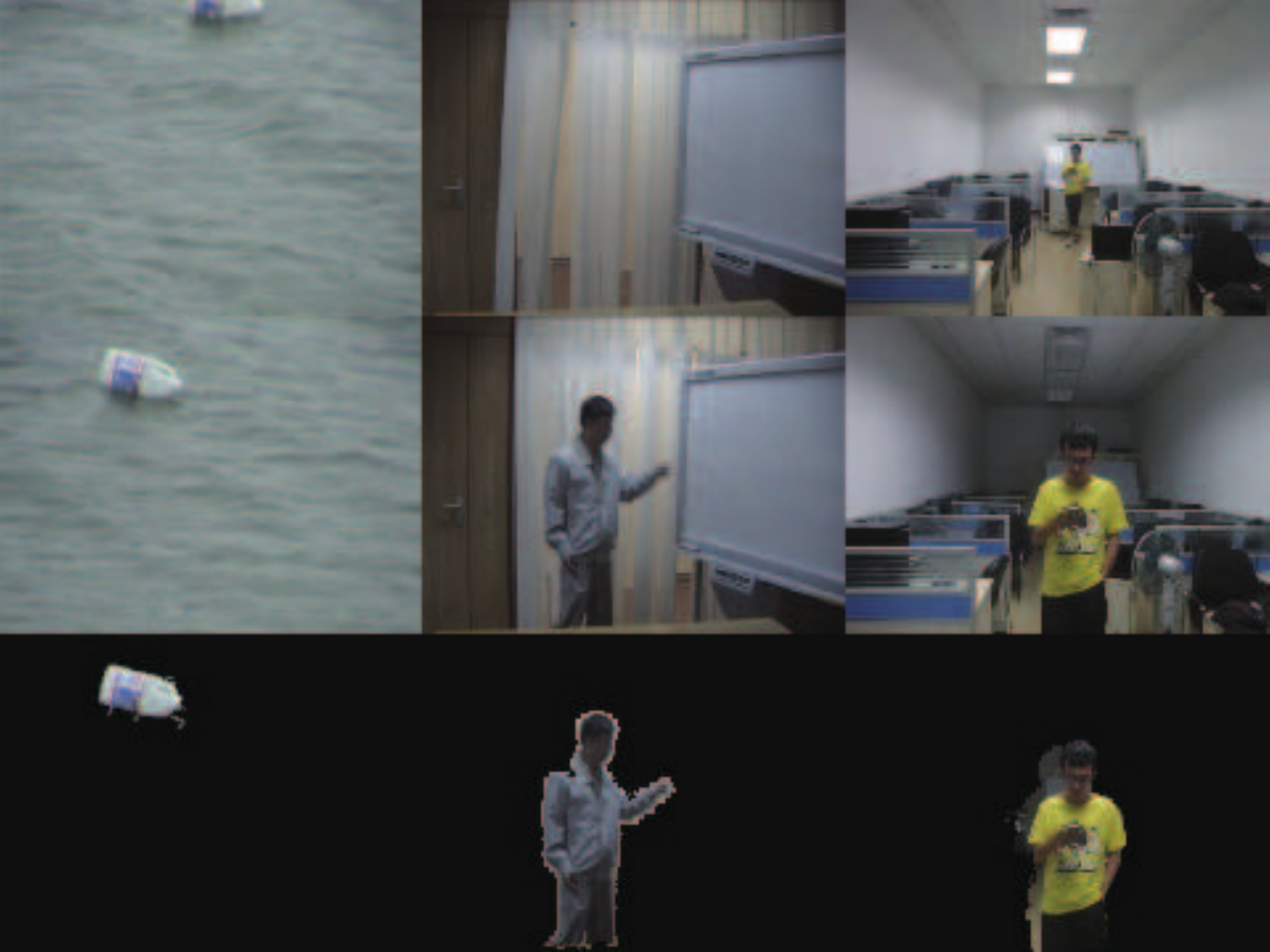}
\end{center}
\caption{Some challenging scenarios for foreground object extraction are handled by our approach: (i) a floating bottle with randomly dynamic water (in the left column), (ii) waving curtains around a person (in the middle column), and (iii) sudden light changing (in the right column).}
\label{fig:exhibit}
\end{figure}

In this paper, we address the above mentioned difficulties by building the background models with the online pursuit of spatio-temporal models. Some results generated by our system for the challenging scenarios are exhibited in Fig.~\ref{fig:exhibit} . Prior to unfolding the proposed approach, we first review the existing works in literature.

\subsection{Related Work}

Due to their pervasiveness in various applications, there is no unique categorization on the existing works of background subtraction. Here we introduce the related methods basically according to their representations, to distinguish with our approach.

The {\bf pixel-processing} approaches modeled observed scenes as a set of independent pixel processes, and they were widely applied in video surveillance applications~\cite{lin2012integrating,liu2011Surveillance} . In these methods~\cite{GGMM,ZGMM,ImGMM,bouwmans2008background}, each pixel in the scene can be described by different parametric distributions (\emph{e.g.} Gaussian Mixture Models) to temporally adapt to the environment changes. The parametric models, however, were not always compatible with real complex data, as they were defined based upon some underlying assumptions. To overcome this problem, some other non-parametric estimations~\cite{NPARA,Bayesian,TIPbayesian08,Vibe} were proposed, and effectively improved the robustness. For example, Barnich et al.~\cite{Vibe} presented a sample-based classification model that maintained a fixed number of samples for each pixel and classified a new observation as background when it matched with a predefined number of samples. Liao et al.~\cite{SILTP} recently employed the kernel density estimation (KDE) technique to capture pixel-level variations. Some distinct scene variations, \emph{i.e.} illumination changes and shadows, can be explicitly alleviated by introducing the extra estimations~\cite{GMMIllu}. Guyon et al.~\cite{guyon2013foreground} proposed to utilize the low rank matrix decomposition for background modeling, where the foreground objects constituted the correlated sparse outliers. Despite acknowledged successes, this category of approaches may have limitations on complex scenarios, as the pixel-wise representations overlooked the spatial correlations between pixels.

The {\bf region-based} methods built background models by taking advantages of inter-pixel relations, demonstrating impressive results on handling dynamic scenes.  A batch of diverse approaches were proposed to model spatial structures of scenes, such as joint distributions of neighboring pixels~\cite{Bayesian,SP10}, block-wise classifiers~\cite{BKClassification}, structured adjacency graphs~\cite{OLISVM}, auto-regression models~\cite{PDTM,Kalman}, random fields~\cite{CRF}, and multi-layer models~\cite{PixelLayers} etc. And a number of fast learning algorithms were discussed to maintain their models online, accounting for environment variations or any structural changes. For example, Monnet et al.~\cite{PDTM} trained and updated the region-based model by the generative subspace learning. Cheng et al.~\cite{OLISVM} employed the generalized 1-SVM algorithm for model learning and foreground prediction. In general, methods in this category separated the spatial and temporal information, and their performances were somewhat limited in some highly dynamic  scenarios, \emph{e.g.} heavy rains or sudden illumination changes.

The third category modeled scene backgrounds by exploiting both spatial and temporal information. Mahadevan et al.~\cite{STSALI} proposed to separate foreground objects from surroundings by judging the distinguished video patches, which contained different motions and appearances compared with the majority of the whole scene. Zhao et al.~\cite{CCIPCABS} addressed the outdoor night background modeling by performing subspace learning within video patches. Spatio-temporal representations were also extensively discussed in other vision tasks such as action recognition~\cite{liang2013action} and trajectory parsing~\cite{liu2013trajectory}.  These methods motivated us to build models upon the spatio-temporal representations, \emph{i.e.} video bricks.

In addition, several {\bf saliency-based} approaches provided alternative ways based on spatio-temporal saliency estimations~\cite{OpFlow,STSALI,BackICCV01}. The moving objects can be extracted according to their salient appearances and/or motions against the scene backgrounds. For example, Wixson et al.~\cite{OpFlow} detected the salient objects according to their consistent moving directions over time. Kim et al.~\cite{STSaliency} used a discriminant center-surround hypothesis to extract foreground objects around their surroundings.

Along with the above mentioned background models, a number of reliable image features were utilized to better handle the background noise~\cite{lin2010layered}. Exemplars included the Local Binary Pattern (LBP) features~\cite{LBP,LTP,CSLBP} and color texture histograms~\cite{ColorText}. The LBP operators described each pixel by the relative graylevels of its neighboring pixels, and their effectiveness has been demonstrated in several vision tasks such as face recognition and object detection~\cite{LBP,LBPMetric,lin2012representing}. The Center-Symmetric LBP was proposed in \cite{CSLBP} to further improve the computational efficiency. Tan and Triggs~\cite{LTP} extended LBP to LTP (Local Ternary Pattern) by thresholding the graylevel differences with a small value, to enhance the effectiveness on flat image regions.

\subsection{Overview}

In this work, we propose to learn and maintain the dynamic models within spatio-temporal video patches (\emph{i.e.} video bricks), accounting for real challenges in surveillance scenarios~\cite{lin2012integrating}. The algorithm can process $15 \sim 20$ frames per second in the resolution $352 \times 288$ (pixels) on average. We briefly overview the proposed framework of background modeling in the following aspects.

{\bf 1. Spatio-temporal representations.} We represent the observed scene by video bricks, \emph{i.e.} video volumes spanning over both spatial and  temporal domain, in order to jointly model spatial and temporal information. Specifically, at every location of the scene, a sequence of video bricks are extracted as the observations, within which we can learn and update the background models. Moreover, to compactly encode the video bricks against illumination variations, we design a brick-based descriptor, namely Center Symmetric Spatio-Temporal Local Ternary Pattern (CS-STLTP), which is inspired by the 2D scale invariant local pattern operator proposed in \cite{SILTP}. Its effectiveness is also validated in the experiments.

{\bf 2. Pursuing dynamic subspaces.} We treat each sequence of video bricks at a certain location as a consecutive signal, and generate the subspace within these video bricks. The linear dynamic system (\emph{i.e.} Auto Regressive Moving Average, ARMA model~\cite{ARMA}) is adopted to characterize the spatio-temporal statistics of the subspace. Specifically, given the observed video bricks, we express them by a data matrix, in which each column contains the feature of a video brick. The basis vectors (\emph{i.e.} eigenvectors) of the matrix can be then estimated analytically, representing the appearance parameters of the subspace, and the parameters of dynamical variations are further computed based on the fixed appearance parameters. It is worth mentioning that our background model jointly captures the information of appearance and motion as the data (\emph{i.e.} features of the video bricks) are extracted over both spatial and temporal domains.

\begin{figure*}[ptb]
\begin{center}
\includegraphics[width=\textwidth]{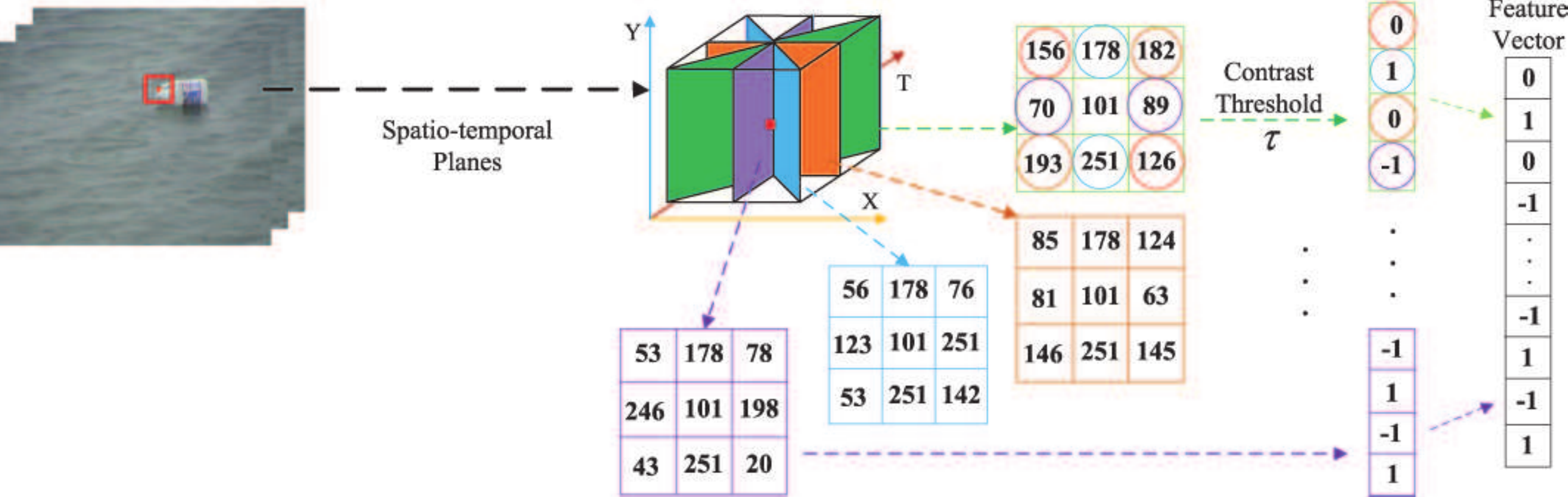}
\end{center}
\caption{An example of computing the CS-STLTP feature. For one pixel in the video brick, we construct four spatio-temporal planes. The center-symmetric local ternary patterns for each plane is calculated, which compares the intensities in a center-symmetric direction with a contrasting threshold $\tau$. The CS-STLTP feature is concatenated by the vectors of the four planes.}
\label{fig:feature}
\end{figure*}

{\bf 3. Maintaining dynamic subspaces online.} Given the newly appearing video bricks with our model, moving foreground objects are segmented by estimating the residuals within the related subspaces of the scene, while the background models are maintained simultaneously to account for scene changes. The raising problem is to update parameters of the subspaces incrementally against disturbance from foreground objects and background noise. The new observation may include noise  pixels (\emph{i.e.} outliers), resulting in degeneration of model updating~\cite{CCIPCABS,PDTM}. Furthermore, one video brick could be partially occluded by foreground objects in our representation, \emph{i.e.} only some of pixels in the brick are true positives. To overcome this problem, we present a novel approach to compensate observations (\emph{i.e.} the observed video bricks) by generating data from the current models. Specifically, we replace the pixels labeled as non-background by the generated pixels to synthesize the new observations. The algorithm for online model updating includes two steps: (i) update appearance parameters using the incremental subspace learning technique, and (ii) update dynamical variation parameters by analytically solving the linear reconstruction. The experiments show that the proposed method effectively improves the robustness during the online processing. 

The remainder of this paper is arranged as follows. We first present the model representation in Section~\ref{sec:represent}, and then discuss the initial learning, foreground segmentation and online updating mechanism in Section~\ref{sec:learn}, respectively. The experiments and comparisons are demonstrated in Section~\ref{sec:exp} and finally comes the conclusion in Section~\ref{sec:conclu} with a summary.

\section{Dynamic Spatio-temporal Model} \label{sec:represent}

In this section, we introduce the background of our model, and then discuss the video brick representation and our model definition, respectively.

\subsection{Background}

In general, a complex surveillance background may include diverse appearances that sometimes move and change dynamically and randomly over time flying~\cite{lin2009stochastic}. There is a branch of works on time-varying texture modeling~\cite{DT,DTRCG,DTSEG} in computer vision. They often treated the scene as a whole, and pursued a global subspace by utilizing the linear dynamic system (LDS). These models worked well on some natural scenes mostly including a few homogeneous textures, as the LDS characterizes the subspace with a set of linearly combined components. However, under real surveillance challenges, it could be intractable to pursue the global subspace. In this work, we represent the observed scene by an array of small and independent subspaces, each of which is defined by the linear system, so that our model is able to handle better challenging scene variations. Our background model can be viewed as a mixed compositional model consisting of the linear subspaces. In particular, we conduct the background subtraction with our model based on the following observations.

{\bf Assumption 1:} The local scene variants (\emph{i.e.} appearance and motion changing over time) can be captured by the low-dimensional subspace.

{\bf Assumption 2:} It is feasible to separate foreground moving objects from the scene background by fully exploiting spatio-temporal statistics.


\subsection{Spatio-temporal Video Brick} \label{sec:feature}

Given the surveillance video of one scene, we first decompose it with a batch of small brick-like volumes. We consider the video brick of small size  (\emph{e.g.}, $4\times 4 \times 5$ pixels) includes relative simple content, which can be thus generated by few bases (components). And the brick volume integrates both spatial and temporal information, that we can better capture complex appearance and motion variations compared with the traditional image patch representations.

We divide each frame $I_i$ , $(i\,=\,1,2,\dots,n)$ into a set of image patches with the width $w$ and height $h$.  A number $t$ of patches at the same location across the frames are combined together to form a brick. In this way, we extract a sequence of video bricks $V = \{v_{1},v_{2},\dots,v_{n}\}$ at every location for the scene.

Moreover, we design a novel descriptor to describe the video brick instead of using RGB values. For any video brick $v_i$, we first apply the CS-STLTP operator on each pixel, and pool all the feature values into a histogram. For a pixel $x_c$, we construct a few 2D spatio-temporal planes centered  at it, and compute the local ternary patterns (LTP) operator~\cite{LTP} on each plane. The CS-STLTP then encodes $x_c$ by combining the LTP operators of all planes. Note that the way of splitting spatio-temporal planes little affects the operator's performance. To simplify the implementation, we make the planes parallel to the Y axis, as Fig.~\ref{fig:feature} shown.

We index the neighborhood pixels of $x$ by $\{0, \ldots, M \}$, the operator response of the $j$-th plane can be then calculated as:

\begin{equation} \label{equ:feature_begin}
\digamma^j(x)\,=\,\overset{\frac{M}{2}-1}{\underset{m=0}{\biguplus}}\,s_{\tau}(p_m,p_{m+\frac{M}{2}}),
\end{equation}
where pixel $k$ and $k+ M/2$ are two symmetric neighbors of pixel $x_c$. $p_k$ and $p_{k+\frac{M}{2}}$ are the graylevels of the two pixels, respectively. The sign $\biguplus$ indicates stretching elements into a vector. The function $s_{\tau}$ is defined as follows:

\begin{equation} \label{equ:feature_end} \begin{aligned}
    s_{\tau}(p_m,p_{m+\frac{M}{2}}) = \left\{ \begin{array}{ll}
                                    1,& \;\;\;\textrm{if $p_m > (1+\tau)p_{m+\frac{M}{2}}$},\\
                                    \text{-}1,& \;\;\;\textrm{if $p_m < (1-\tau)p_{m+\frac{M}{2}}$},\\
                                    0,& \;\;\;\textrm{otherwise}.
                                    \end{array} \right.
\end{aligned} \end{equation}
where $\tau$ is a constant threshold for the comparing range.

Suppose that we take $M=8$ neighborhood pixels for computing the operator in each spatio-temporal plane, and the number of planes is $4$. The resulting CS-STLTP vector contains $M/2 \times 4 = 16$ bins. Fig.\ref{fig:feature} illustrates an example of computing the CS-STLTP operator, where we apply the operator for one pixel on $4$ spatial-temporal planes displayed with different colors (\emph{e.g.}, green, blue, purple and orange).

Then we build a histogram for each video brick by accumulating the CS-STLTP responses of all pixels. This definition was previously proposed by Guo et al~\cite{LBPMetric}.

\begin{equation}
H(k) = \Sigma_{x \in v_i} \Sigma_{j = 1}^4 \bf{1}( \digamma^j(x), k ), ~~k \in [0, K ],
\end{equation}
where $\bf{1}(a, b)$ is an indicator function, i.e. $\bf{1}(a, b) = 1$ only if $a = b$. To measure the operator response, we transform the binary vector of CS-STLTP into a uniform value that is defined as the number of spatial transitions (bitwise changes) following, as discussed in~\cite{LBPMetric} .  For example, the pattern (\emph{i.e.} the vector of $16$ bins) $0000000000000000$ has a value of $0$ and $1000000000000000$ of $1$. In our implementation, we further quantize all possible values into $48$ levels. To further improve the capability, we can generate histograms in each color channel and concatenate them together.

The proposed descriptor is computationally efficient and compact to describe the video brick. In addition, by introducing a tolerative comparing range in the LTP operator computation, it is robust to local spatio-temporal noise within a range.

\subsection{Model Definition} \label{sec:model}

Let $m$ be the descriptor length for each brick, and $V = \{v_1, v_2, \dots, v_n\}$, $v_i \in \mathbb{R}^{m}$ be a sequence of video bricks at a certain location of the observed background. We can use a set of bases (components) $\mathbf{C} = [C_1,C_2,\dots,C_d]$ to represent the subspace where $V$ lies in. Each video brick $v_i$ in $V$ can be represented as

\begin{equation}
    v_i = \underset{j=1}{\overset{d}{\sum}} \, z_{i,j} C_j + \omega_i,
\label{eq:videobrick}
\end{equation}
where $C_j$ is the $j$-th basis ($j$-th column of matrix $\mathbf{C}$) of the subspace, $z_{i,j}$ the coefficient for $C_j$, and $\omega_i$ the appearance residual. We denote $\mathbf{C}$ to represent appearance consistency of the sequence of video bricks. In some traditional background models by subspace learning, $z_{i,j}$ can be solved and kept as a constant, with the underlying assumption that the appearance of background would be stable within the observations. In contrast, we treat $z_{i,j}$ as the variable term that can be further phrased as the time-varying state, accounting for temporally coherent variations (\emph{i.e.} the motions). For notation simplicity, we neglect the subscript $j$, and denote $Z = \{ z_1, z_2, \ldots, z_n \}$ for all the bricks. The dynamic model is formulated as,

\begin{equation}\label{eq:varying}
    z_{i+1}\,=\,Az_i \,+\, \eta_i,
\end{equation}
where $\eta_i$ is the state residual, and $A$ is a matrix of $ d \times d$ dimensions to model the variations. With this definition, we consider $A$ representing the temporal coherence among the observations.

Therefore, the problem of pursuing dynamic subspace is posed as solving the appearance consistency $\mathbf{C}$ and the temporal coherence $A$, within the observations. Since the sequence states $Z$ are unknown, we shall jointly solve $\mathbf{C}$, $A$, $Z$ by minimizing an empirical energy function $\mathcal{F}_n(\mathbf{C},A,Z)$:

\begin{equation} \label{equ:formulation}
    \min\, \mathcal{F}_n(\mathbf{C},A,Z) = \frac{1}{2n}\,\underset{i=1}{\overset{n}{\sum}} \, \| v_i - Cz_i \|_2^2 + \| z_i - A z_{i-1} \|_2^2 .
 \end{equation}
Here $\mathcal{F}_n(\mathbf{C},A,Z)$ is not completely convex but we can solve it by fixing either $Z$ or $(\mathbf{C}, A)$. Nevertheless, its computation cost is expensive for learning the entire background online. Here we simplify the dynamic model in Equation (\ref{eq:varying}) into a linear system, following the auto-regressive moving average  (ARMA) process. In literature, Soatto et al.~\cite{DT} originally associated the output of ARMA model with dynamic textures, and showed that the first-order ARMA model, driven by white zero-mean Gaussian noise, can capture a wide range of dynamic textures. In our approach, the difficulty of modeling the dynamic variations can be alleviated due to the brick-based representation, i.e. the observed scene is decomposed into video bricks. Thus, we consider the ARMA process a suitable solution to model the time-varying variables, which can be solved efficiently. Specifically, we introduce a robustness term (\emph{i.e.} matrix) $B$, which includes a number $d_\epsilon$ of bases, and we set $\eta_i = B \epsilon_i$, where $\epsilon_i$ denotes the noise.

We further summarize the proposed dynamic model, where we add the subscript $n$ to the main components, indicating they are solved within a number $n$ of observations, as,

\begin{equation} \label{equ:model} \begin{aligned}
    & v_i\,=\,\mathbf{C}_{n}\,z_i\,+\,\omega_{i}, \\
    & z_{i+1}\,=\,A_{n}\,z_{i}\,+\,B_{n}\,\epsilon_{i}, \\
    & \omega_{i}\,\overset{IID}{\sim}\,N(0,\Sigma_{\omega}), \;\;\epsilon_{i}\,\overset{IID}{\sim}\,N(0,I_{d_\epsilon}).
\end{aligned} \end{equation}
In this model,  $\mathbf{C}_n \in \mathbb{R}^{m \times d}$ and $A_n \in \mathbb{R}^{d \times d}$ represent the appearance consistency and temporal coherence, respectively. $B_{n} \in \mathbb{R}^{d \times d_{\epsilon}} $ is the robustness term constraining the evolution of $Z$ over time.  $\omega_i \in \mathbb{R}^{m}$ indicates the residual corresponding to observation $v_i$, and $\epsilon_{i} \in \mathbb{R}^{d_\epsilon}$ the noise of state variations.  During the subspace learning, $\omega_i$ and $\epsilon_{i}$ are assumed to follow the zero-mean Gaussian distributions. Given a new brick mapped into the subspace, $\omega_i$ and $\epsilon_{i}$ can be used to measure how likely the observation is suitable with the subspace, so that we utilize them for foreground object detection during online processing.

The proposed model is time-varying, and the parameters $\mathbf{C}_n, A_n, B_n$ can be updated incrementally along with the processing of new observations, in order to adapt our model with scene changes.




\section{Learning Algorithm}\label{sec:learn}

In this section, we discuss the learning for spatio-temporal background models, including initial subspace generation and online maintenance. The initial learning is performed at the beginning of system deployment, when only a few foreground objects move in the scene. Afterwards, the system switches to the mode of online maintenance. 

\subsection{Initial Model Learning}\label{sec:offllearn}

In the initial stage, the model defined in Equation (\ref{equ:model}) can be degenerated as a non-dynamic linear system, as the $n$ observations are extracted and fixed. Given a brick sequence $V = \{v_{1},v_{2},\dots,v_{n}\}$, we present an algorithm to identify the model parameters $\mathbf{C}_{n},\,A_{n},\,B_{n}$, following the sub-optimal solution proposed in~\cite{DT}.

To guarantee the Equation (\ref{equ:model}) has an unique and canonical solution, we postulate

\begin{equation} \label{equ:assumption} \begin{aligned}
    &n \gg d,\;\;\text{Rank}(\mathbf{C}_n)\,=\,d,\;\;\mathbf{C}_n^\top \mathbf{C}_n\,=\,I_d,
\end{aligned} \end{equation}
where $I_d$ is the identity matrix of dimension $d \times d$. The appearance consistency term $\mathbf{C}_n$ can be estimated as,

\begin{equation} \label{equ:learnC} \begin{aligned}
    \mathbf{C}_n\,&=\,\underset{\mathbf{C}_n}{\operatorname{arg\,min}}\;\vert\; W_n - \mathbf{C}_{n}\,[\;z_1\;z_2\;\cdots\;z_n\;]\; \vert \\
\end{aligned} \end{equation}
where $W_n$ is the data matrix composed of observed video bricks $[v_1,v_2,\cdots,v_n]$. The equation.(\ref{equ:learnC}) satisfies the full rank approximation property and can be thus solved by the singular value decomposition (SVD). We have,

\begin{eqnarray}
W_n = U \Sigma Q^\top, \\\nonumber
U^\top U = I, Q^\top Q = I,
\end{eqnarray}
where $Q$ is the unitary matrix, $U$ includes the eigenvectors, and $\Sigma$ is the diagonal matrix of the singular values. Thus, $\mathbf{C}_{n}$ is treated as the first $d$ components of $U$, and the state matrix $[z_1\;z_2\;\cdots\;z_n]$ as the product of $d \times d$ sub-matrix of $\Sigma$ and the first $d$ columns of $Q^\top$.

The temporal coherence term $A_{n}$ is calculated by solving the following linear problem:

\begin{equation}
    A_{n}\,=\,\underset{A_n}{\operatorname{arg\,min}}\; \vert \; [\;z_2\;z_3\;\cdots\;z_{n}\;] - A_{n}[\;z_1\;z_2\;\cdots\;z_{n-1}\;] \; \vert.
\end{equation}
The statistical robustness term $B_{n}$ is estimated by the reconstruction error $E$

\begin{equation} \begin{aligned}
    E & \,=\,[\;z_2\;z_3\;\cdots\;z_{n}\;]\,-\,A_{n}\,[\;z_1\;z_2\;\cdots\;z_{n-1}\;] \\
             & \,=\,B_{n}\,[\;\epsilon_1\;\epsilon_2\;\cdots\;\epsilon_{n-1}\;],
\end{aligned} \end{equation}
where $B_{n} \cong \frac{1}{\sqrt{n-1}}\,E$. Since the rank of $A_n$ is $d$ and $d \ll n$, the rank of input-to-state noise $d_\epsilon$ is assumed to be much smaller than $d$. That is, the dimension of $E$ can be further reduced by SVD: $E = U_\epsilon\,\Sigma_\epsilon\,Q_\epsilon^\top$, and we have

\begin{equation} \begin{aligned}
    & B_{n}\!=\!\frac{1}{\sqrt{n-1}}\left[\!\begin{array}{cccc} U_\epsilon^1 & \cdots & U_\epsilon^{d_\epsilon} \end{array}\!\right]\!\left[\! \begin{array}{cccc} \Sigma_\epsilon^1 & & \\
    & \ddots & \\
    & & \Sigma_\epsilon^{d_\epsilon} \end{array}\!\right].
\end{aligned} \end{equation}

The values of $d$, $d_\epsilon$ essentially imply the complexity of subspace from the aspects of appearance consistence and temporal coherence, respectively. For example, video bricks containing static content can be well described with a function of low dimensions while highly dynamic video bricks (\emph{e.g.}, from an active fountain) require more bases to generate. In real surveillance scenarios, it is not practical to pre-determine the complexity of scene environments. Hence, in the proposed method, we adaptively determine $d$, $d_\epsilon$ by thresholding eigenvalues in $\Sigma$ and $\Sigma_\epsilon$, respectively.

\begin{equation} \label{equ:dyndim} \begin{aligned}
    & d^\ast\,=\,\underset{d}{\operatorname{arg\,max}}\;\Sigma^d\,>\,T_d, \\
    & d_\epsilon^\ast\,=\,\underset{d_\epsilon}{\operatorname{arg\,max}}\;\Sigma_\epsilon^{d_\epsilon}\,>\,T_{d_\epsilon},
\end{aligned} \end{equation}
where $\Sigma^d$ indicates the $d$-th eigenvalue in $\Sigma$ and $\Sigma_\epsilon^{d_\epsilon}$ the $d_\epsilon$-th eigenvalue in $\Sigma_\epsilon$.

\subsection{Online Model Maintenance}\label{sec:ollearn}

Then we discuss the online processing with our model that segments foreground moving objects and keeps the model updated.

{\bf (I) Foreground segmentation.~~} Given one newly appearing video brick $v_{n+1}$, we can determine whether pixels in $v_{n+1}$ belong to the background or not by thresholding their appearance residual and state residual. We first estimate the state of $v_{n+1}$ with the existing $\mathbf{C}_{n}$,

\begin{equation} \label{equ:seg_begin}
    z'_{n+1}\,=\,\mathbf{C}_{n}^\top \, v_{n+1},
\end{equation}
and further the appearance residual of $v_{n+1}$

\begin{equation}\label{equ:appearance}
    \omega_{n+1}\,=\,v_{n+1}\,-\,\mathbf{C}_n z'_{n+1}.
\end{equation}

As the state $ z_n$ and the temporal coherence $A_n$ have been solved, we can then estimate the state residual $\epsilon_n$ according to Equation (\ref{equ:model}),

\begin{equation} \label{equ:seg_end} \begin{aligned}
    & B_n \epsilon_{n} \,=\, z'_{n+1} \,-\, A_n z_n \\
    \Rightarrow\, & \epsilon_{n} \,=\, \operatorname{pinv}(B_n)\,(z'_{n+1} \,-\, A_n z_n),
\end{aligned} \end{equation}
where $\operatorname{pinv}$ denotes the operator of pseudo-inverse.

With the state residual $\epsilon_{n}$ and the appearance residual $\omega_{n+1}$ for the new video brick $v_{n+1}$, we conduct the following criteria for foreground segmentation, in which two thresholds are introduced.
\begin{enumerate}
      \item $v_{n+1}$ is classified into background, only if all dimensions of $\epsilon_{n}$ are less than a threshold $T_\epsilon$.
      \item If $v_{n+1}$ has been labeled as non-background, perform the pixel-wise segmentation by comparing $\omega_{n+1}$ with a threshold $T_\omega$: the pixel is segmented as foreground if its corresponding dimension in $\omega_{n+1}$ is greater than $T_\omega$.
\end{enumerate}

{\bf (II) Model updating.~~} During the online processing, the key problem for model updating is to deal with foreground disturbance, \emph{i.e.} to avoid absorbing pixels from foreground objects or noise.

In this work, we develop an effective approach to update the model with the synthesized data. We first generate a video brick from the current model, namely noise-free brick, $\hat{v}_{n+1}$, as

\begin{equation} \label{equ:Syn} \begin{aligned}
    & \hat{z}_{n+1}\,=\,A_{n}z_{n}, \\
    & \hat{v}_{n+1}\,=\,\mathbf{C}_{n}\,\hat{z}_{n+1}.
\end{aligned} \end{equation}
Then we extract pixels from $\hat{v}_{n+1}$ to compensate occluded (\emph{i.e.} foreground) pixels in the newly appearing brick. Concretely, the pixels labeled as non-background are replaced by the pixels from the noise-free video brick at the same place. We can thus obtain a synthesized video brick $\bar{v}_{n+1}$ for model updating.

Given the brick $\bar{v}_{n+1}$, the data matrix $W_{n}$ composed of observed video bricks is extended to $W_{n+1}$. Then we update the model $\mathbf{C}_{n+1}$ according to Equation (\ref{equ:learnC}).

Our algorithm of model updating includes two steps: (i) update parameters for appearance consistency $\mathbf{C}_{n+1}$ by employing the incremental subspace learning technique, and (ii) update parameters of state variations $A_{n+1}$, $B_{n+1}$.

{\bf (i) Step 1.~~}  For the $d$-dimension subspace, with eigenvectors $\mathbf{C}_n$ and eigenvalues $\Lambda_n$, its covariance matrix $\operatorname{Cov}_n$ can be approximated as

\begin{equation} \label{equ:Basis}
    \operatorname{Cov}_n \,\thickapprox\, \overset{d}{\underset{j=1}{\sum}}\,\lambda_{n,j} c_{n,j} c_{n,j}^\top \,=\, \mathbf{C}_n \Lambda_n \mathbf{C}_n^\top,
\end{equation}
where $c_{n,j}$ and $\lambda_{n,j}$ denote the $j$-th eigenvector and eigenvalue, respectively. With the newly synthesized data $\bar{v}_{n+1}$, the updated covariance matrix $\operatorname{Cov}_{n+1}$ is formulated as

\begin{equation} \begin{aligned}
    \operatorname{Cov}_{n+1} \,&=\,(1-\alpha)\operatorname{Cov}_n\,+\,\alpha\,\bar{v}_{n+1}\,\bar{v}_{n+1}^\top \\
                &\approx\,(1-\alpha)\,\mathbf{C}_n \Lambda_n \mathbf{C}_n^\top\,+\,\alpha\,\bar{v}_{n+1}\,\bar{v}_{n+1}^\top \\
                &=\,\overset{d}{\underset{i=1}{\sum}}\,(1-\alpha)\,\lambda_{n,i}\,c_{n,i}\,c_{n,i}^\top + \alpha\,\bar{v}_{n+1}\,\bar{v}_{n+1}^\top,
\end{aligned} \end{equation}
where $\alpha$ denotes the learning rate. The covariance matrix can be further re-formulated to simplify computation, as,

\begin{equation}
    \operatorname{Cov}_{n+1} \,=\, Y_{n+1} Y_{n+1}^\top,
\end{equation}
where $Y_{n+1} = [y_{n+1,1}\,y_{n+1,2}\,\dots\,y_{n+1,d+1}]$ and each column $y_{n+1,j}$ in $Y_{n+1}$ is defined as

\begin{equation}
    y_{i} = \left\{ \begin{array}{ll}
        \sqrt{1-\alpha\lambda_j}\,c_{n,i}, & \;\;\;\text{if } 1 < j < d, \\
        \sqrt{\alpha}\,\bar{v}_{n+1}, & \;\;\;\text{if } j = d+1. \end{array} \right.
\end{equation}

To reduce the computation cost, we can estimate $\mathbf{C}_{n+1}$ by a smaller matrix $Y_{n+1}^\top Y_{n+1}$, instead of the original large matrix $\operatorname{Cov}_{n+1}$.

\begin{equation} \label{equ:eigen_decom}
    (Y_{n+1}^\top\,Y_{n+1})\,e_{n+1,j} \,=\, \lambda_{n+1,j}\,e_{n+1,j} \quad j = 1,2,\dots,d+1,
\end{equation}
where $e_{n+1,j}$ and $\lambda_{n+1,j}$ are the $j$-th eigenvector and eigenvalue of matrix $Y_{n+1}^\top Y_{n+1}$, respectively. Let $c_{n+1,j} = Y_{n+1} e_{n+1,j}$, and we re-write Equation (\ref{equ:eigen_decom}) as

\begin{equation} \label{equ:RIPCA_end} \begin{aligned}
    & Y_{n+1}\,Y_{n+1}^\top\,Y_{n+1}\,e_{n+1,j} \,=\, \lambda_{n+1,j}\,Y_{n+1}\,e_{n+1,j}, \\
    & \operatorname{Cov}_{n+1}\,c_{n+1,j} \,=\, \lambda_{n+1,i}\,c_{n+1,j}\;\;\;\;\;\;j = 1,2,\dots,d+1.
\end{aligned} \end{equation}
We thus obtain the updated eigenvectors $\mathbf{C}_{n+1}$ and the corresponding eigenvalues $\Lambda_{n+1}$ of the new covariance matrix $\operatorname{Cov}_{n+1}$. Note that the dimension of the subspace is automatically increased along with the newly added data $\bar{v}_{n+1}$. To guarantee the appearance parameters remain stable, we keep the main principal (\emph{i.e.} top $d$) eigenvectors and eigenvalues while discarding the least significant components.

The above incremental subspace learning algorithm has been widely applied in several vision tasks such as face recognition and image segmentation~\cite{IPCA2,IPCA3,IPCA4}, and also for background modeling in \cite{CCIPCABS,INSUBL,ICA09}. However, the noise observations caused by moving objects or scene variations often disturb the subspace maintenance, \emph{e.g.} the eigenvectors could change dramatically during the processing. Many efforts~\cite{RPCA11}\cite{ding2011bayesian} have been dedicated to improve the robustness of incremental learning by using statistical  analysis. Several discriminative learning algorithms~\cite{diana2012background} were also employed to train background classifiers that can be incrementally updated. In this work, we utilize a version of Robust Incremental PCA (RIPCA)~\cite{IPCA1} to cope with the outliers in $\bar{v}_{n+1}$. Note that $\bar{v}_{n+1}$ consists of pixels either from the generated data $\hat{v}_{n+1}$ or real videos, where outliers may exist in some dimensions.

In the traditional PCA learning, the solution is derived by minimizing a least-squared reconstruction error,

\begin{equation}
    \min |r_{n+1}|^2\,=\, | \mathbf{C}_{n}\mathbf{C}_{n}^\top \bar{v}_{n+1} - \bar{v}_{n+1} |^2. \label{equ:error}
\end{equation}
Following \cite{IPCA1}, we impose a robustness function $w(t) = \frac{1}{1+(t/\rho)^2}$ over each dimension of $r_{n+1}$, and the target can be re-defined as,

\begin{equation}
    \min \sum_j (r^k_{n+1})^2\, \leftarrow \, w(r_{n+1}^k) (r^k_{n+1})^2,
\end{equation}
where the superscript $k$ indicates the $k$-th dimension. The parameter $\rho$ in the robustness function is estimated by

\begin{equation} \label{equ:IRPCA_begin} \begin{aligned}
    & \rho\,=\,[\rho^1, \rho^2, \dots, \rho^{|\bar{v}_{n+1}|}]^\top\\
    & \rho^k \,=\,\underset{i=1}{\overset{d}{\max}}\,\beta\sqrt{\lambda_{n,i}}\,|\,c_{n,j}^k\,|,\; j = 1, 2, \dots, |\bar{v}_{n+1}|
\end{aligned} \end{equation}
where $\beta$ is a fixed coefficient. The $k$-th dimension of $\rho$ is proportional to the maximal projection of the current eigenvectors on the $k$-th dimension, (\emph{i.e.} $\rho^k$ is weighted by their corresponding eigenvalues). Note that $w(r_{n+1}^k)$ is a function of the residual error which should be calculated for each vector dimension. And the computation cost for $w(r_{n+1}^k)$ can be neglected in the analytical solution. 


Accordingly, we can update the observation $\bar{v}_{n+1}$ over each dimension by computing the function $w(r_{n+1}^k)$, 

\begin{equation}
\tilde{v}^k_{n+1}\,=\,\sqrt{w(r^k_{n+1})}\,\bar{v}^k_{n+1}.
\label{equ:robust}
\end{equation}
That is, we treat $\tilde{v}_{n+1}$ as the new observation during the procedure of incremental learning.

{\bf (i) Step 2.~~} With the fixed $\mathbf{C}_{n+1}$, we then update the parameters of state variations $A_{n+1}$, $B_{n+1}$. We first estimate the latest state $z_{n+1}$ based on the updated $\mathbf{C}_{n+1}$ as,

\begin{equation} \label{equ:updateAB_begin}
    z_{n+1} = \mathbf{C}^{\top}_{n+1} \tilde{v}_{n+1}.
\end{equation}
$A_{n+1}$ can be further calculated, by re-solving the linear problem of a fixed number of latest observed states,

\begin{equation} \label{equ:updateAB_end} \begin{aligned}
    A_{n+1}\,[\;z_{n-l+1}\;\cdots\;z_{n}\;]\,=\,[\;z_{n-l+2}\;\cdots\;z_{n+1}\;],
\end{aligned} \end{equation}
where $l$ indicates the number of latest observed states, \emph{i.e.} the span of observations. And similarly, we update $B_{n+1}$ by computing the new reconstruction error $E=[z_{n-l+2}\;\cdots\;z_{n+1}]-A_{n+1}\,[z_{n-l+1}\;\cdots\;z_{n}]$.

\begin{figure}[ptb]
\begin{center}
\includegraphics[width=3.2in]{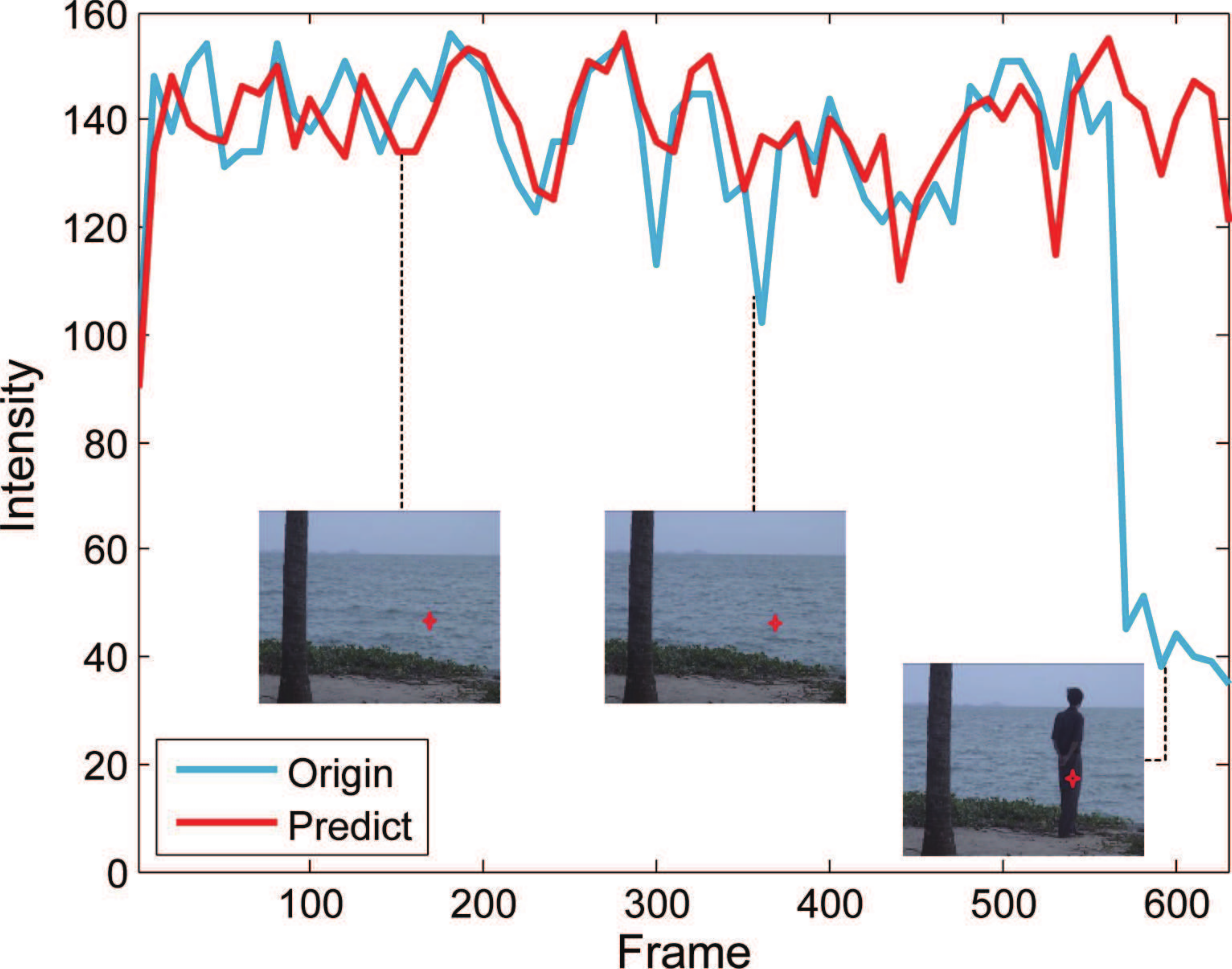}
\end{center}
\caption{An example to demonstrate the robustness of model maintenance. In the scenario of dynamic water surfaces, we visualize the original and predicted intensities for a fixed position (denoted by the red star), with the blue and red curves, respectively. With our updating scheme, when the position is occluded by a foreground object during from frame $551$ to $632$, the predicted intensities are not disturbed by foreground, \emph{i.e.} the model remains stable.}
\label{fig:robust}
\end{figure}

\begin{algorithm}[ptb]
\KwIn{ Video brick sequence $V = \{v_{1},v_{2},\dots,v_{n}\}$ for every location for the scene. }
\KwOut{Maintained Background models and foreground regions}
\ForAll{locations for the scene}
{

    Given the observed video bricks $V$, extract the CS-STLTP descriptor\;

     Initialize the subspace by estimating $\mathbf{C}_{n},A_{n},B_{n}$  using Equation (\ref{equ:assumption})-(\ref{equ:dyndim})\;
    \For{the newly appearing video brick $v_{n+1}$}
    {
        (1) Extract the CS-STLTP descriptor for $v_{n+1}$\;
        (2) Calculate its state residual $\epsilon_{n}$ and appearances residual $\omega_{n+1}$ by Equation (\ref{equ:appearance}) and (\ref{equ:seg_end})\;
        (3) For each pixel of $v_{n+1}$, classify it into foreground or background by thresholding the two residuals with $\epsilon_{n}$, $\omega_{n+1}$\;
        (4) Generate the noise-free brick $\hat{v}_{n+1}$ from the current model by Equation (\ref{equ:Syn})\;
        (5) Synthesize video brick $\bar{v}_{n+1}$ for model updating\;
        (6) Update $\bar{v}_{n+1}$ into $\tilde{v}_{n+1}$ by introducing a robustness function\;
        (7) Update the new appearance parameter $\mathbf{C}_{n+1}$ by calculating the covariance matrix $\operatorname{Cov}_{n+1}$ with the learning rate $\alpha$\;
        (8) Update the state variation parameters $A_{n+1},\,B_{n+1}$ \;
    }
}
\caption{The sketch of the proposed algorithm.}\label{alg:STDS}
\end{algorithm}

We present an empirical study in Fig.~\ref{fig:robust} to demonstrate the effectiveness of this updating method. The video for background modeling includes dynamic water surfaces. Here we visualize the original and predicted intensities for a fixed position (denoted by the red star), with the blue and red curves, respectively. We can observe that the model remains stable against foreground occlusion.

{\em Time complexity analysis.~~} We mainly employ SVD and linear programming in the initial learning. The time complexity of SVD is $O(n^3)$ and the learning time of linear programming is $O(n^2)$. For a certain location, the time complexity of initial learning is $O(n^3) + O(n^2) = O(n^3)$ for each subspace, where $n$ denotes the number of video bricks for model learning. As for online learning, incremental subspace learning and linear programming are utilized. Given a $d$-dimension subspace, the time complexity for component updating (\emph{i.e.} step $1$ of the model maintenance) is $O(dn^2)$. Thus, the total time complexity for online learning is $O(dn^2) + O(l^2)$, where $l$ is the number of states used to solve the linear problem.

We summarize the algorithm sketch of our framework in Algorithm~\ref{alg:STDS}.

\section{Experiments}\label{sec:exp}

In this section, we first introduce the datasets used in the experiments and the parameter settings, then present the experimental results and comparisons. The discussions of system components are proposed at last.

\subsection{Datasets and settings}

We collect a number of challenging videos to validate our approach, which are publicly available or from real surveillance systems. Two of them (AirportHall and TrainStation) from the PETS database\footnote{Downloaded from http://www.cvg.rdg.ac.uk/slides/pets.html.} include crowded pedestrians and moving cast shadows; five highly dynamic scenes \footnote{Downloaded from http://perception.i2r.a-star.edu.sg} include waving curtain active fountain, swaying trees, water surface; the others contain extremely difficult cases such as heavy rain, sudden and gradual light changing. Most of the videos include thousands of frames, and some of the frames are manually annotated as the  ground-truth provided by the original databases.

Our algorithm has been adopted in a real video surveillance system and achieves satisfactory performances. The system is capable of processing $15 \sim 20$ frames per second in the resolution $352 \times 288$ pixels. The hardware architecture is an Intel i7 2600 (3.4 GHz) CPU and 8GB RAM desktop computer.

All parameters are fixed in the experiments, including the contrast threshold for CS-STLTP descriptor $\tau=0.2$, the dimension threshold for ARMA model  $T_d = 0.5$, $T_{d_\epsilon} = 0.5$, the span of observations for model updating $l = 60$, and the size of bricks $4\times 4\times 5$. For foreground segmentation, the threshold of appearance residual $T_\omega = 3$, update threshold $T_\epsilon = 3$ and $T_\omega = 5$, $T_\epsilon = 4$ for RGB. In the online model maintenance, the coefficient $\beta = 2.3849$, the learning rate $\alpha = 0.05$ for RIPCA. 

In the experiments, we use the first $50$ frames of each testing video to initialize our system (i.e. to perform the initial learning), and keep model updated in the rest of sequence.  In addition, we utilize a standard post-processing to eliminate areas including less than $20$ pixels. All other competing approaches are executed with the same setting as our approach. 

We utilize the F-score as the benchmark metric, which measures the segmentation accuracy by considering both the recall and the precision. The F-score is defined as

\begin{equation}
  F\,=\,\frac{2\,TP}{2\,TP\,+\,FP\,+\,FN},
\end{equation}
where TP is true positives (foreground objects), FN false negatives (false background pixels), FP false positive (false foreground pixels).

\begin{figure}[ptb]
\begin{center}
\includegraphics[width=90mm]{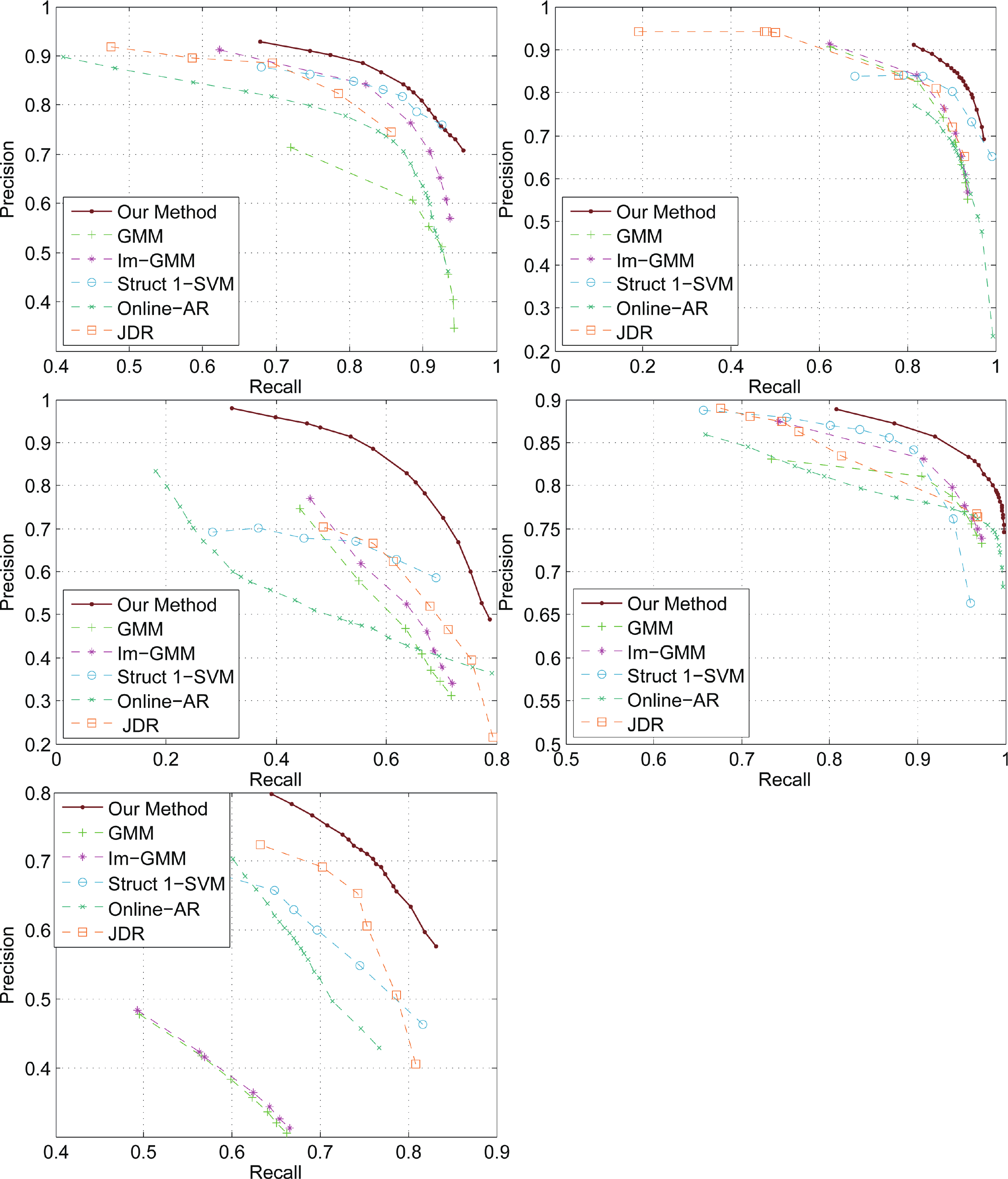}
\end{center}
\caption{Experimental results generated by our approach and competing methods on $5$ videos: first row left, the scene including a dynamic curtain and indistinctive foreground objects (\emph{i.e.} having similar appearance with backgrounds); first row right, the scene with heavy rain; second row left, an indoor scene with the sudden lighting changes; second row right, the scene with dynamic water surface; third row, a busy airport. The precision-recall (PR) curve is introduced as the benchmark measurement for all the $6$ algorithms.}\label{fig:PR}
\end{figure}

\subsection{Experimental results}

\begin{small}
\begin{small}
\begin{table*}[ptb]
    \center
    \caption{Quantitative results and comparisons on the $10$ complex videos using the F-score ($\%$) measurement. The last two columns report the results of our method using either RGB or CS-STLTP as the feature.}
    \begin{tabular}{ccccccccc}
    \addlinespace
    \toprule
    Scene & \tabincell{c}{GMM\cite{GGMM}} & \tabincell{c}{Im-GMM\cite{ZGMM}} & \tabincell{c}{Online-AR\cite{PDTM}} & \tabincell{c}{JDR\cite{Bayesian}} & \tabincell{c}{SVM\cite{OLISVM}} & \tabincell{c}{PKDE\cite{SILTP}} & \tabincell{c}{STDM(RGB)} & \tabincell{c}{STDM(Ftr.)} \\
    \midrule
    \multicolumn{1}{l}{1\# Airport} & 46.99 & 47.36 & 62.72 & 60.23 & 65.35 & 68.14 & \textbf{70.52} & 66.40 \\
    \multicolumn{1}{l}{2\# Floating Bottle} & 57.91 & 57.77 &  43.79 & 45.64  & 47.87 & 59.57 & 69.04 & \textbf{78.17} \\
    \multicolumn{1}{l}{3\# Waving Curtain} & 62.75 & 74.58 & 77.86 & 72.72 & 77.34 & 78.01 & \textbf{79.74} & 74.93 \\
    \multicolumn{1}{l}{4\# Active Fountain} & 52.77 & 60.11 & 70.41 & 68.53 & 74.94 & 76.33 & 76.85 & \textbf{85.46} \\
    \multicolumn{1}{l}{5\# Heavy Rain} & 71.11 & 81.54 & 78.68 & 75.88 & \textbf{82.62} & 76.71 &   79.35 & 75.29 \\
    \multicolumn{1}{l}{6\# Sudden Light} & 47.11 & 51.37 & 37.30  & 52.26 & 47.61 & 52.63 & 51.56 &  \textbf{74.57} \\
    \multicolumn{1}{l}{7\# Gradual Light} & 51.10 & 50.12 & 13.16 & 47.48 & 62.44 & 54.86 & 54.84	&  \textbf{77.41} \\
    \multicolumn{1}{l}{8\# Train Station} & 65.12 & 68.80  & 36.01 & 57.68 & 61.79 & 67.05 & \textbf{73.43} &  66.35 \\
    \multicolumn{1}{l}{9\# Swaying Trees} & 19.51 & 23.25 & 63.54 & 45.61 & 24.38 & 42.54 & 43.71 &  \textbf{75.89} \\
    \multicolumn{1}{l}{10\# Water Surface} & 79.54 & 86.01 & 77.31 & 84.27 & 83.13 & 74.30  & 88.54 &  \textbf{88.68} \\
    \multicolumn{1}{l}{Average} & 55.39 & 59.56 & 57.02 & 60.23 & 59.79 & 63.08 & 68.75 & \textbf{76.31} \\
    \bottomrule
    \end{tabular}
  \label{tab:F-score}
\end{table*}
\end{small}
\end{small}

\begin{figure*}[ptb]
\begin{center}
\includegraphics[width=140mm]{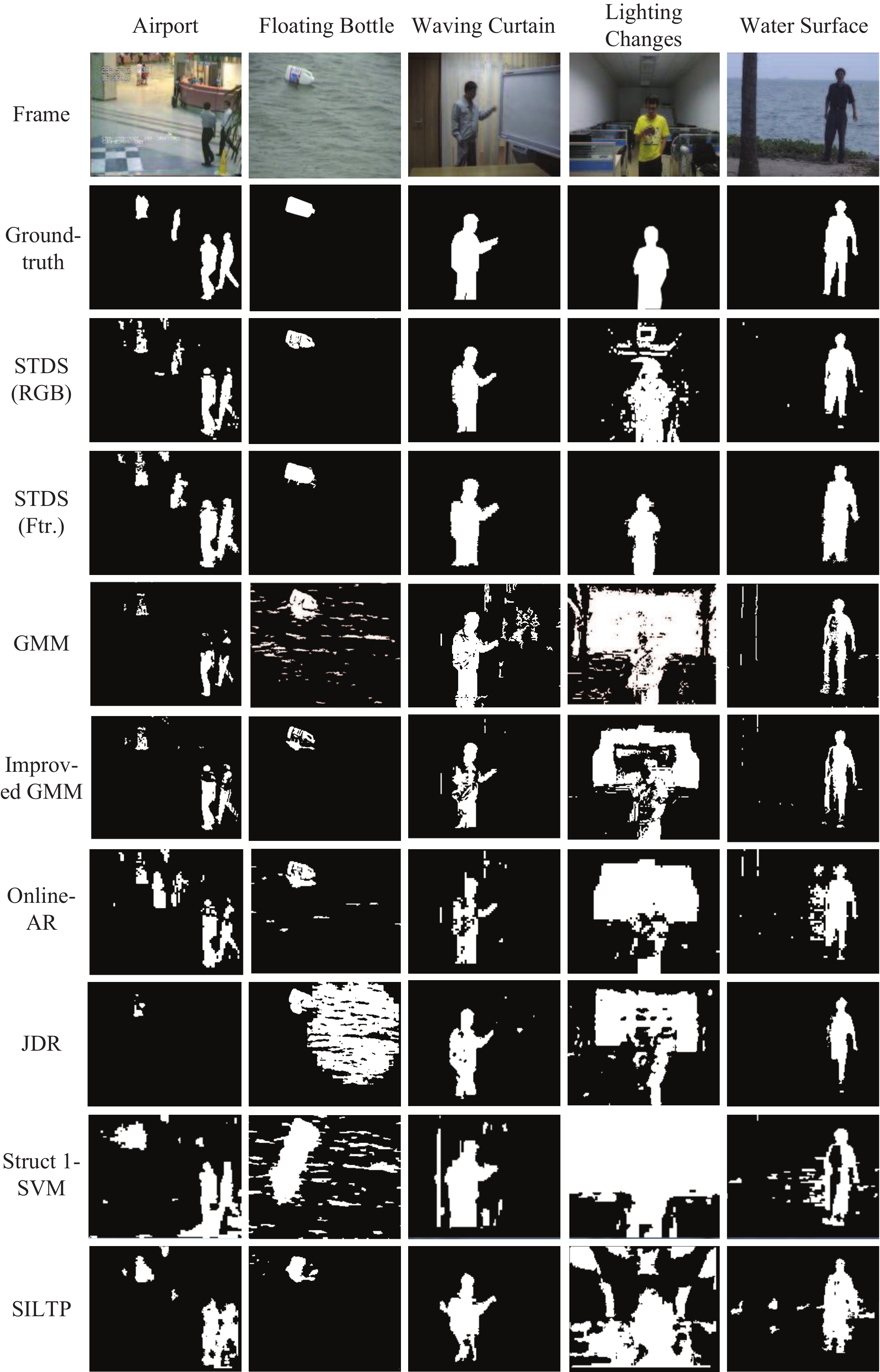}
\end{center}
\caption{Sampled results of background subtraction generated by our approach (using RGB or CS-STLTP as the feature and RIPCA as the update strategy) and other competing methods.}
\label{fig:results_backsub}
\end{figure*}

{\em Experimental results}. We compare the proposed method (STDM) with six state-of-the-art online background subtraction algorithms including Gaussian Mixture Model (GMM)~\cite{GGMM} as baseline, improved GMM~\cite{ZGMM}\footnote{Available at http://dparks.wikidot.com/background-subtraction}, online auto-regression model~\cite{PDTM}, non-parametric model with scale-invariant local patterns~\cite{SILTP}, discriminative model using generalized Struct 1-SVM~\cite{OLISVM}\footnote{Available at http://www.cs.mun.ca/$\sim$gong/Pages/Research.html}, and the Bayesian joint domain-range (JDR) model~\cite{Bayesian}\footnote{Available at http://www.cs.cmu.edu/$\sim$yaser/}. In the comparisons, for the methods~\cite{GGMM,ZGMM,OLISVM,Bayesian} we use their released codes, and implement the methods~\cite{SILTP,PDTM} by ourselves. The F-scores ($\%$) over all $10$ videos are reported in Table~\ref{tab:F-score}, where the last two columns report results of our method using either RGB or CS-STLTP as the feature. Note that for the result using the RGB feature we represent each video brick by concatenating the RGB values of all its pixels. We also exhibit the results and comparisons using the precision-recall (PR) curves, as shown in Fig.~\ref{fig:PR}. Due to space limitation, we only show results on $5$ videos. From the results, we can observe that the proposed method outperforms the other methods in most videos in general. For the scenes with highly dynamic backgrounds (\emph{e.g.}, the $\# 2$ $\#5$ and $\# 10$ scenes), the improvements made by our method are more than $10\%$. And the system enables us to well handle the indistinctive foreground objects (\emph{i.e.} small objects or background-like objects in the $\# 1$, $\# 3$ scenes). Moreover, we make significant improvements (\emph{i.e.} $15\% \sim 25\%$) in the scene \#6 and \#7 including both sudden and gradual lighting changes. A number of sampled results of background subtraction are exhibited in Fig.~\ref{fig:results_backsub}.

The benefit of using the proposed CS-STLTP feature is clearly validated by observing the results shown in Table~\ref{tab:F-score} and Fig.~\ref{fig:results_backsub}.  In general, our approach simply using RGB values can achieve satisfying performances for the common scenes, \emph{e.g.}, with fair appearance and motion changes, while the CS-SILTP operator can better handle highly dynamic variations (\emph{e.g.} sudden illumination changing, rippling water).  In addition, we also compare CS-STLTP with the existing scale invariant descriptor SILTP proposed in \cite{SILTP}. We reserve all settings in our approach except replacing the feature by SILTP, and achieve the average precision over all $10$ videos: $69.70\%$. This result shows that CS-STLTP is very suitable and effective for the video brick representation.

\subsection{Discussion}\label{sec:exp_dis}

Furthermore, we conduct the following empirical studies to justify the parameter determinations and settings of our approach.

\begin{figure}[ptb]
\begin{center}
\includegraphics[width=40mm]{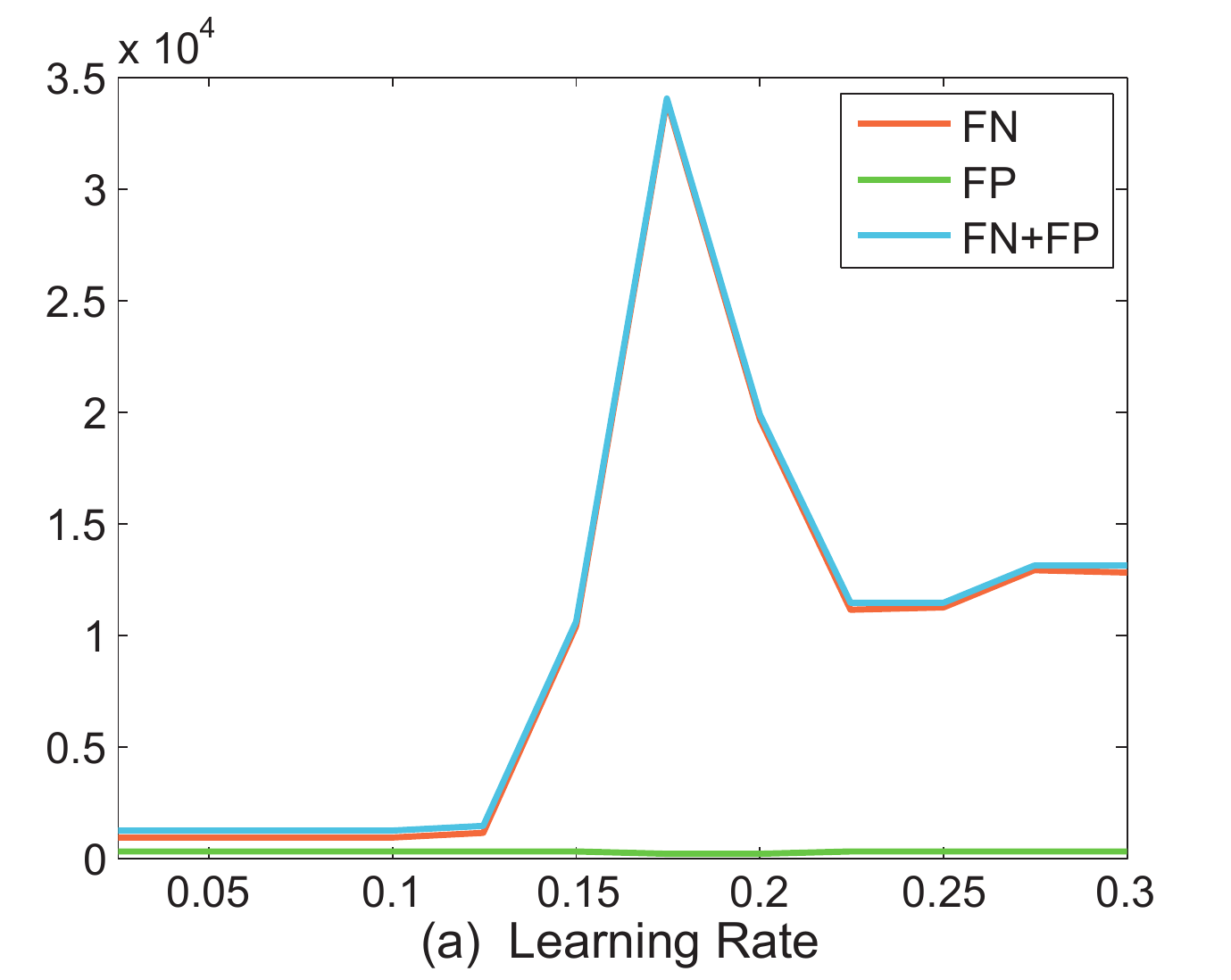}
\includegraphics[width=40mm]{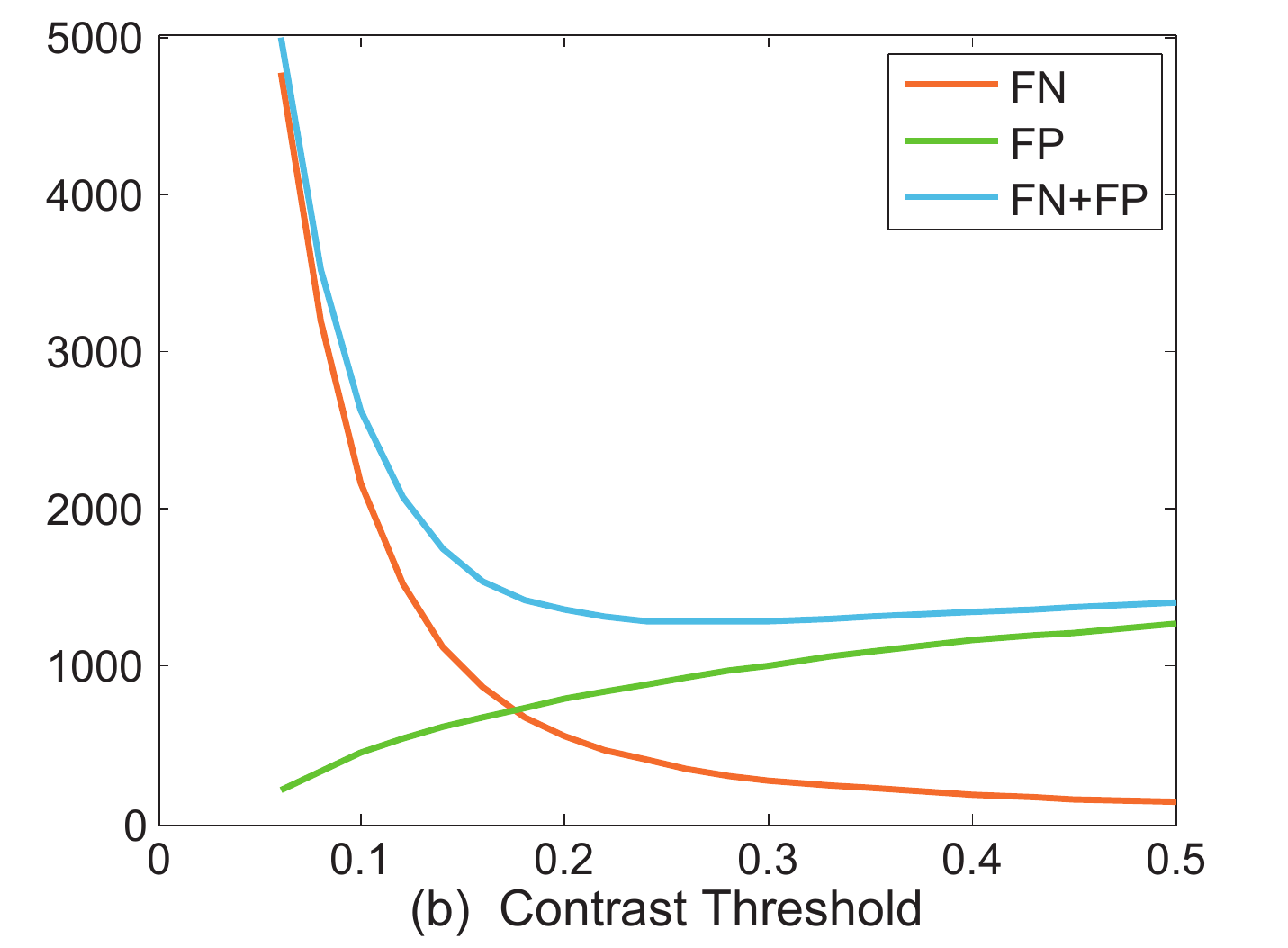}
\end{center}
\caption{Discussion of parameter selection: (i) learning rate $\alpha$ for model maintenance (in (a)) and (ii) the contrast threshold of CS-STLTP feature $\tau$ (in (b)). In each figure, the horizontal axis represents the different parameter values; the three lines in different colors denote, respectively, the false positive (FP), false negative (FN), and the sum of FP and FN.}\label{fig:discuss_para}
\end{figure}

\begin{figure}[ptb]
\begin{center}
\includegraphics[width=2.6in]{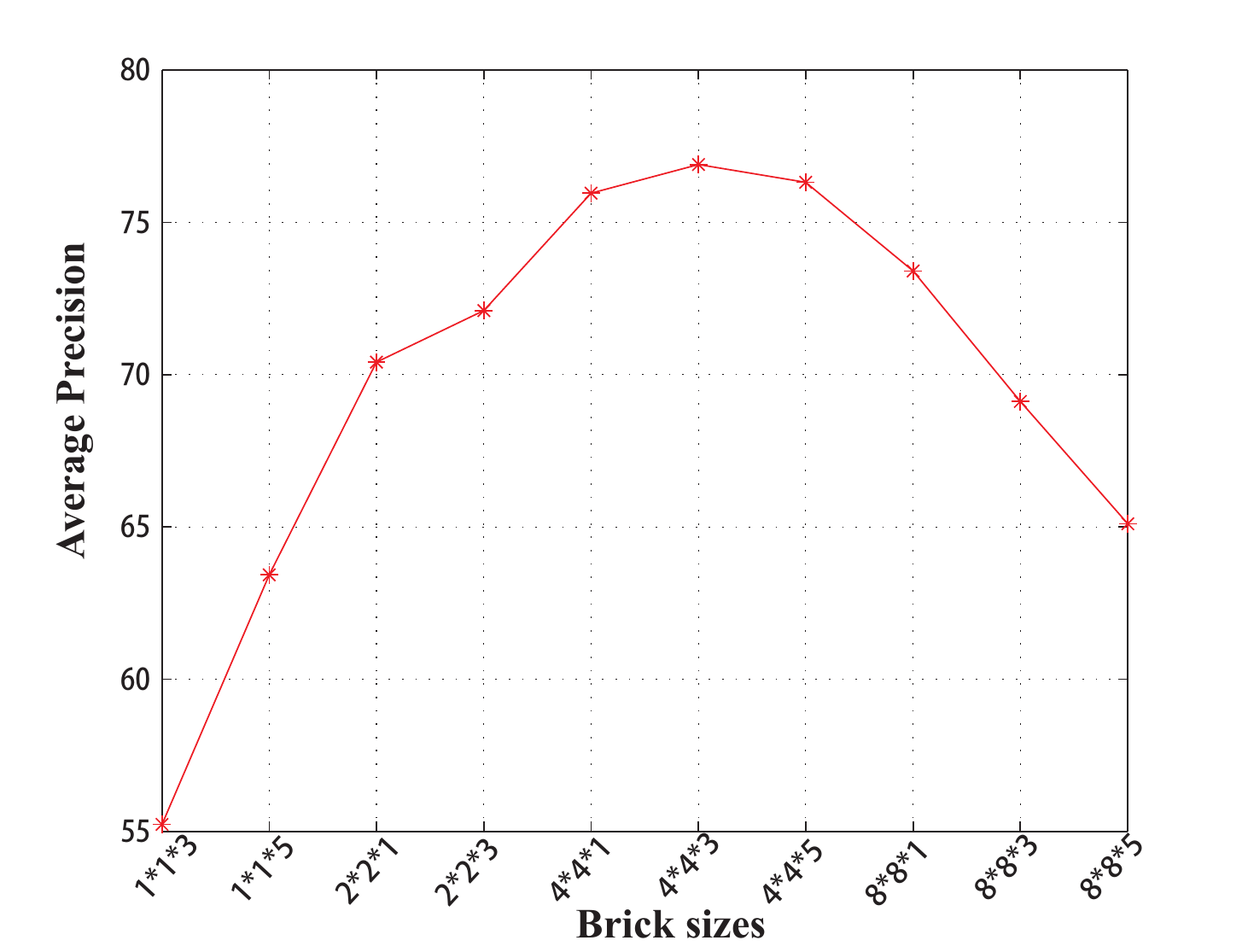}
\caption{Empirical study for the size of video brick in our approach. We carry on the experiments on the $10$ videos with different brick size while keeping the rest settings. The vertical axis represents the average precisions of background subtraction and the horizontal represents the different sizes of video bricks with respect to background decomposition.}\label{fig:bricksize}
\end{center}
\end{figure}

{\em Efficiency.}  Like other online-learning background models, there is a trade-off between the model stability and maintenance efficiency. The corresponding parameter in our method is the learning rate $\alpha$. We tune $\alpha$ in the range of $0 \sim 0.3$ by fixing the other model parameters and visualize the quantitative results of background subtraction, as shown in Fig.~\ref{fig:discuss_para}(a). From the results, we can observe this parameter is insensitive in range $0 \sim 0.1$ in our model. In practice, once the scene is extremely busy and crowded, it could be set as a relative small value to keep the model stable.

{\em Feature effectiveness.} The contrast threshold $\tau$ is the only parameter in CS-STLTP operator, which affects the power of feature to character spatio-temporal information within video bricks. From the empirical results of parameter tuning, as shown in Fig.~\ref{fig:discuss_para} (b), we can observe that the appropriate range for $\tau$ is $0.15 \sim 0.25$. In practice, the model could become sensitive to noise by setting a very small value of $\tau$ (say $\tau < 0.15$), and too large $\tau$ (say $\tau > 0.25$) might reduce the accuracy on detecting foreground regions with homogeneous appearances.


{\em Size of video brick.}  One may be interested in how the system performance is affected by the size of video brick for background decomposition, so that we present an empirical study on different sizes of video bricks in Fig. \ref{fig:bricksize}. We observe that the best result is achieved with the certain brick size of $4 \times 4 \times 3$, and the results with the sizes of  $4 \times 4 \times 1$ and $4 \times 4 \times 5$ are also satisfied.  As of very small bricks (\emph{e.g.} $1 \times 1 \times 3$ ), few spatio-temporal statistics are captured and the models may have problems on handling scene variations.  The bricks of large sizes (\emph{e.g.} $8 \times 8 \times 5$ ) carry too much information, and their subspaces cannot be effectively generated by the linear ARMA model. The experimental results are also accordant with our motivations in Section I.  In practice, we can flexibly set the size according to the resolutions of surveillance videos.

{\em Model initialization.} Our method is not sensitive to the number of observed frames in the initial stage of subspace generation. We test the different numbers, say $30$, $40$, $60$, on two typical surveillance scenes, i.e. the Airport Hall (scene $\# 1$) and the Train Station (scene $\# 8$). The F-score outputs show the deviations with different numbers of initial frames are very small, e.g. less than $0.2$. In general, we require the observed scenes to be relatively clean for initialization, although a few  objects that move across are allowed.


\section{Conclusion}\label{sec:conclu}

This paper studies an effective method for background subtraction, addressing the all challenges in real surveillance scenarios. In the method, we learn and maintain the dynamic texture models within spatio-temporal video patches (\emph{i.e.} video bricks). Sufficient experiments as well as empirical analysis are presented to validate the advantages of our method.

In the future, we plan to improve the method in two aspects. (1) Some efficient tracking algorithms can be employed into the framework to better distinguish the foreground objects. (2) The GPU-based implementation can be developed to process each part of the scene in parallel, and it would probably significantly improve the system efficiency.

\bibliographystyle{unsrt}
\bibliography{mybib}

\begin{IEEEbiography}[{\includegraphics[width=1in,height=1.25in,clip,keepaspectratio]{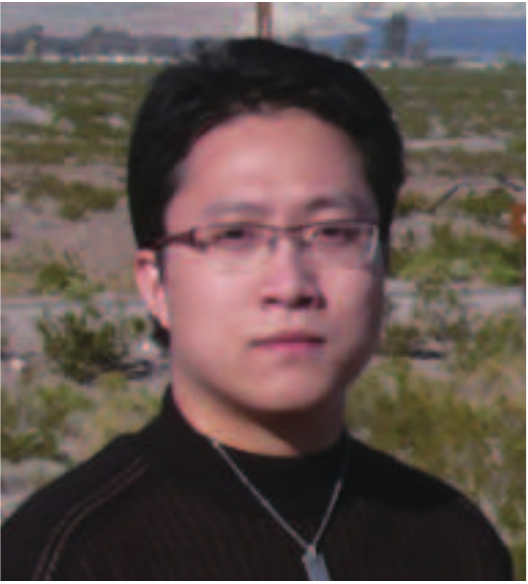}}]{Liang Lin} is a full Professor with the School of Advanced Computing, Sun Yat-Sen University (SYSU), China. He received the B.S. and Ph.D. degrees from the Beijing Institute of Technology (BIT), Beijing, China, in 1999 and 2008, respectively. From 2006 to 2007, he was a joint Ph.D. student with the Department of Statistics, University of California, Los Angeles (UCLA). His Ph.D. dissertation was achieved the China National Excellent PhD Thesis Award Nomination in 2010. He was a Post-Doctoral Research Fellow with the Center for Vision, Cognition, Learning, and Art of UCLA. His research focuses on new models, algorithms and systems for intelligent processing and understanding of visual data such as images and videos. He has published more than 50 papers in top tier academic journals and conferences including Proceedings of the IEEE, T-PAMI, T-IP, T-CSVT, T-MM, Pattern Recognition, CVPR, ICCV, ECCV, ACM MM and NIPS. He was supported by several promotive programs or funds for his works, such as the ``Program for New Century Excellent Talents'' of Ministry of Education (China) in 2012, the ``Program of Guangzhou Zhujiang Star of Science and Technology'' in 2012, and the Guangdong Natural Science Funds for Distinguished Young Scholars in 2013. He received the Best Paper Runners-Up Award in ACM NPAR 2010, and Google Faculty Award in 2012. 
\end{IEEEbiography}

\begin{IEEEbiography}[{\includegraphics[width=1in,height=1.25in,clip,keepaspectratio]{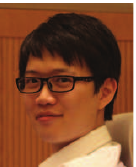}}]{Yuanlu Xu} has just finished his master studies at School of Information Science and Technology, Sun Yat-sen University and is going to pursue a Ph.D. degree at UCLA. His current advisor is Prof. Liang Lin and they have cooperated on publishing a couple of papers on computer vision. Yuanlu received a BE degree with honors from the School of Software, Sun Yat-sen University. His research interests are in video surveillance, image matching and statistical modeling and inference.
\end{IEEEbiography}

\begin{IEEEbiography}[{\includegraphics[width=1in,height=1.25in,clip,keepaspectratio]{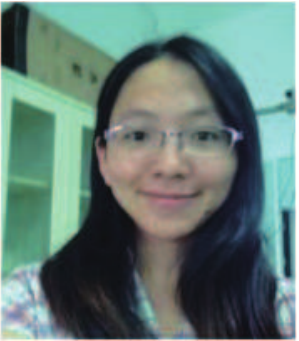}}]{Xiaodan Liang}
received the B.B.A degree in the School of Software, Sun Yat-Sen University in 2010. She is now a Ph.D candidate with the School of Information Science and Technology, Sun Yat-Sen University (SYSU), China. She has published several research papers in top tier academic conferences  and journals.  Her research focuses on structured vision models and multimedia understanding. 
\end{IEEEbiography}

\begin{IEEEbiography}[{\includegraphics[width=1in,height=1.25in,clip,keepaspectratio]{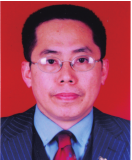}}]{Jianhuang Lai}received his M.Sc. degree in applied mathematics in 1989 and his Ph.D. in mathematics in 1999 from SUN YAT-SEN University, China. He joined Sun Yat-sen University in 1989 as an Assistant Professor, where currently, he is a Professor with the Department of Automation of School of Information Science and Technology and dean of School of Information Science and Technology. His current research interests are in the areas of digital image processing, pattern recognition, multimedia communication, wavelet and its applications. He has published over 100 scientific papers in the international journals and conferences on image processing and pattern recognition��e.g. IEEE TPAMI , IEEE TNN, IEEE TIP, IEEE TSMC (Part B), Pattern Recognition, ICCV, CVPR and ICDM. Prof. Lai serves as a standing member of the Image and Graphics Association of China and also serves as a standing director of the Image and Graphics Association of Guangdong. 
\end{IEEEbiography}

\end{document}